\documentclass[10pt,twocolumn,letterpaper]{article}

\pdfoutput=1
\usepackage{alex}
\usepackage{iccv}
\usepackage{times}
\usepackage{epsfig}
\usepackage{graphicx}
\usepackage{amsmath}
\usepackage{amssymb}

\usepackage{booktabs} 
\usepackage{subcaption} 
\usepackage{algorithm} 
\usepackage{algpseudocode} 
\usepackage{rotating}
\usepackage{enumitem}
\usepackage[font={small}]{caption}

\usepackage{pgfplots}
\usepackage{tango}
\usepackage{tikz}
\usetikzlibrary{calc,positioning,arrows}
\usetikzlibrary{automata,positioning}
\usetikzlibrary{fit} 
\usetikzlibrary{backgrounds} 
\usetikzlibrary{positioning}
\usetikzlibrary{shadows}
\usetikzlibrary{bayesnet}
\usepgflibrary{shapes}
\tikzstyle{observed}=[circle, thick, minimum size=0.9cm, draw=black!100, fill=black!20]
\tikzstyle{void}=[circle, thick, minimum size=0.9cm, draw=black!0, fill=black!0]
\tikzstyle{lstm}=[rectangle, minimum size=0.2cm, draw=black!100]
\tikzstyle{shadeplate}=[rectangle, thick, inner sep=0.4cm, draw=black!100]
\tikzstyle{table}=[circle,fill=blue!20,draw=black!100,inner sep=1pt, minimum size=30pt]
\tikzstyle{client}=[rectangle,fill=blue!20,draw=black!100,inner sep=1pt, minimum size=12pt]

\usepgfplotslibrary{colorbrewer}
\definecolor{Dark2-8-1}{RGB}{27,158,119}
\definecolor{Dark2-8-A}{RGB}{27,158,119}
\definecolor{Dark2-8-2}{RGB}{217,95,2}
\definecolor{Dark2-8-B}{RGB}{217,95,2}
\definecolor{Dark2-8-3}{RGB}{117,112,179}
\definecolor{Dark2-8-C}{RGB}{117,112,179}
\definecolor{Dark2-8-4}{RGB}{231,41,138}
\definecolor{Dark2-8-D}{RGB}{231,41,138}
\definecolor{Dark2-8-5}{RGB}{102,166,30}
\definecolor{Dark2-8-E}{RGB}{102,166,30}
\definecolor{Dark2-8-6}{RGB}{230,171,2}
\definecolor{Dark2-8-F}{RGB}{230,171,2}
\definecolor{Dark2-8-7}{RGB}{166,118,29}
\definecolor{Dark2-8-G}{RGB}{166,118,29}
\definecolor{Dark2-8-8}{RGB}{102,102,102}
\definecolor{Dark2-8-H}{RGB}{102,102,102}

\usepackage[pagebackref=true,breaklinks=true,letterpaper=true,bookmarks=false]{hyperref}
\hypersetup{
    colorlinks=true,
    linkcolor=black,
    citecolor=black,
}
\pgfplotsset{compat = 1.3}

\newcommand{\lowp}{{\scriptsize \colorbox{Dark2-8-1}{\textcolor{white}L}}\ }
\newcommand{\upp}{{\scriptsize \colorbox{Dark2-8-2}{\textcolor{white}U}}\ }
\newcommand{\outp}{{\scriptsize \colorbox{Dark2-8-3}{\textcolor{white}O}}\ }
\newcommand{\hosp}{{\scriptsize \colorbox{Dark2-8-4}{\textcolor{white}H}}\ }
\newcommand{\glp}{{\scriptsize \colorbox{Dark2-8-5}{\textcolor{white}G}}\ }

\iccvfinalcopy 

\newcommand{\cc}{\textcolor{black}} 
\newcommand{\CHECK}{\textcolor{black}} 

\ificcvfinal\fi
\begin{document}

\title{Learning the Latent ``Look": Unsupervised Discovery of a\\Style-Coherent Embedding from Fashion Images}

\author{Wei-Lin Hsiao\\
UT-Austin\\
{\tt\small kimhsiao@cs.utexas.edu}
\and
Kristen Grauman\\
UT-Austin\\
{\tt\small grauman@cs.utexas.edu}
}

\maketitle

\begin{abstract}
What defines a visual style?
Fashion styles emerge organically from how people assemble outfits of clothing,
making them difficult to pin down with a computational model.
Low-level visual similarity can be too specific to detect stylistically similar images, while manually crafted style categories can be too abstract to capture subtle style differences.
We propose an unsupervised approach to learn a style-coherent representation.  Our method leverages probabilistic polylingual topic models based on visual attributes to discover a set of latent style factors.
Given a collection of unlabeled fashion images, our approach mines for the latent styles, then summarizes outfits by how they mix those styles.  Our approach can organize galleries of outfits by style without requiring any style labels.  Experiments on over 100K images demonstrate its promise for retrieving, mixing, and summarizing fashion images by their style.

\end{abstract}


\section{Introduction}
Computer vision methods that can understand fashion could transform how individual consumers shop for their clothing as well as how the fashion industry can analyze its own trends at scale.  The scope for impact is high: fashion is already a \$1.2 trillion USD global industry, popular social commerce websites like Chictopia and Polyvore draw millions of users, and online subscription services like StitchFix blend algorithms and stylists to personalize shopping selections.  In sync with these growing possibilities, recent research explores new vision methods for fashion, with exciting advances for parsing clothing~\cite{yamaguchi-cvpr2012,paperdoll-iccv2013}, recognizing clothing, attributes~\cite{gallagher-eccv2012,mixmatch2015}, and styles~\cite{hipster,128floats},
matching clothing seen on the street to catalogs~\cite{where-to-buy-iccv2015,street-to-shop2012,getting-the-look2013,runway-realway,deepfashion}, and recommending clothing~\cite{magic-closet,mcauley,dyadic,davis,iwata}.



\begin{table}[t]
  \centering
  \small
  \setlength{\tabcolsep}{0.3em} 
  \ra{1.1}
  \begin{tabular}{@{}cccc@{}}
   \multirow{2}{*}{Query} & Instance match & Label-based & Latent looks \\
    &{\footnotesize -- \textcolor{scarletred3}{low diversity}. }&{\footnotesize -- \textcolor{scarletred3}{inconsistent}. }&{\footnotesize -- \textcolor{chameleon3}{consistent}, \textcolor{chameleon3}{diverse}.} \\
    {\includegraphics[width=0.11\textwidth]{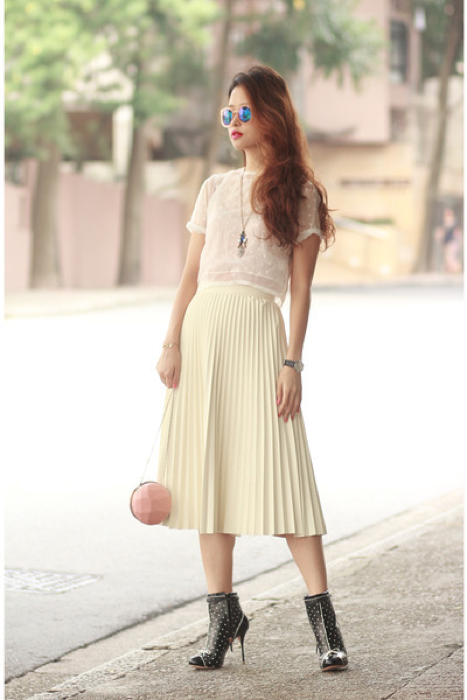}} &
    {\includegraphics[width=0.105\textwidth]{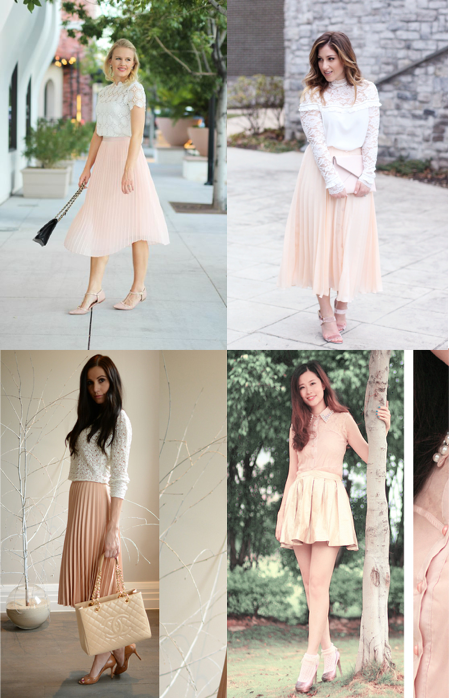}} &
    {\includegraphics[width=0.11\textwidth]{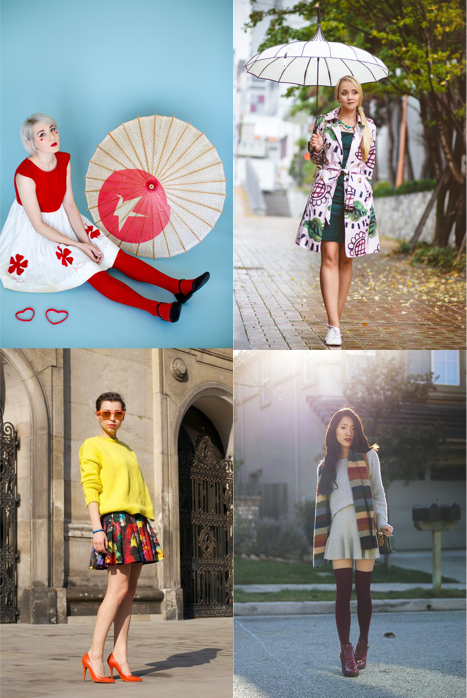}} &
    {\hspace{-0.45em} \includegraphics[width=0.105\textwidth]{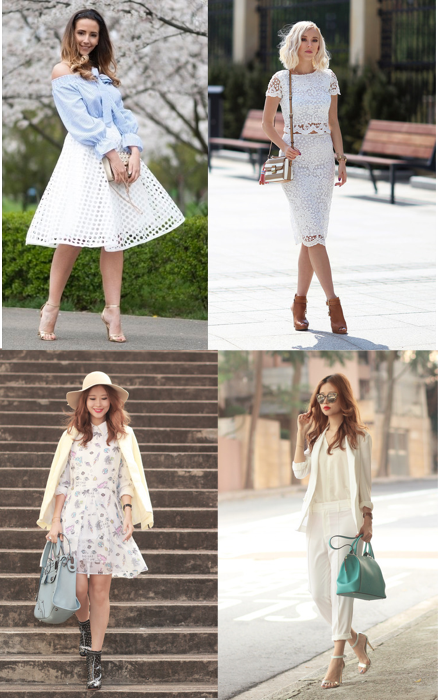}}\\
    & (a) & (b) & (c)
\end{tabular}
\vspace*{-0.1in}
\caption*{Figure 1: The leftmost query image relates to instances like those in (a),(b),(c) in distinct ways.  
In contrast to matching near-duplicate outfits (a) or classifying broad styles (b), we propose to
discover the latent ``looks''---compositions of clothing elements that are stylistically similar (c).}
\label{fig:concept}
\end{table}

Despite substantial progress on all these fronts, capturing the \emph{style-coherent similarity} between outfits remains an important challenge.  In particular, a visual representation with style coherency would capture the \emph{relationship between clothing outfits that share a ``look", even though they may differ in their specific composition of garments.}
A style-coherent representation would be valuable for i) browsing, where a consumer wants to peruse diverse outfits similar in style, ii) recommendation, where a system should suggest new items that add novelty to a consumer's closet without straying from his/her personal style, and iii) style trend tracking, where analysts would like to understand the popularity of items over time.

Style coherency differs from traditional notions of visual similarity.
The problem of style coherency sits between the two extremes currently studied in the literature: on one end of the spectrum are methods that seek robust \emph{instance matching}, e.g., to allow a photo of a garment seen on the street to be matched to a catalog~\cite{where-to-buy-iccv2015,street-to-shop2012,getting-the-look2013,runway-realway,deepfashion} (see Figure~1(a)); on the other end of the spectrum are methods that seek \emph{coarse style classification}, e.g., to label an outfit as one of a small number predefined categories like Hipster, Preppy, or Goth~\cite{hipster,128floats} (see Figure~1(b)).  In contrast to these two extremes, \emph{style coherency} refers to consistent fine-grained trends exhibited by distinct assemblies of garments.  In other words, coherent styles reflect some latent ``look".  See Figure~1(c).

We propose an unsupervised approach to learn a style-coherent representation.  Given a large repository of unlabeled fashion images, the goal is to discover the latent factors that naturally guide how people dress---that is, the underlying compositions of visual attributes that define styles.  To this end, we explore probabilistic topic models.  Topic models in natural language processing~\cite{blei2003latent,rosen2004author} represent text documents as distributions over concepts, each of which is a distribution over words.  In our case, an outfit is a ``document", a predicted visual attribute (e.g., \emph{polka dotted}, \emph{flowing}, \emph{wool}) is a ``word", and each style is a discovered ``topic".  Furthermore, we consider \emph{polylingual} topic models~\cite{mimno2009polylingual} in order to model style consistency across multiple regions of the body, enforcing that the discovered style factors for each region of the body should interact compatibly.  Our idea makes it possible to organize galleries of outfits by style without any style labels.  Building on the proposed model, we develop methods to \emph{mix} styles or \emph{summarize} an image gallery by its styles.

The proposed approach is well-suited for the problem at hand, for several reasons.  First, being unsupervised, our algorithm \emph{discovers} the underlying elements of style, as opposed to us manually defining them.  It is often difficult to manually craft labels, especially in the domain of fashion.  Clothing styles emerge organically from instances of what people choose to wear---not from some top-down pre-ordained bins of outfit types---and furthermore they evolve over time, meaning today's hand-crafted lexicon will eventually fade in relevance.  Secondly, our approach accounts for the fact that style is about the Gestalt: individual items do not dictate a style; rather, it is their \emph{composition} that creates a look~\cite{barthes2013language}.  Finally, our topic model approach also naturally accounts for the soft boundaries of style, describing outfits as mixtures of overlapping styles.

Our experiments on two challenging fashion datasets~\cite{hipster,deepfashion} demonstrate the advantages of our unsupervised representation \CHECK{compared to style-based CNNs and more basic attribute descriptions}.
We validate that our styles align well with human perceived styles.  We further show their value for retrieval, mixing, and summarization.  As a secondary contribution, we introduce a new dataset of 19K images labeled for fine-grained, body-localized attributes relevant for fashion analysis.

\vspace*{-0.1in}
\section{Related Work}

\paragraph{Attributes for fashion}

Describable attributes, such as \emph{floral}, \emph{denim}, \emph{long-sleeved}, are often of interest in analyzing fashion images. 
Prior work explores a variety of recognition schemes~\cite{di2013style,mixmatch2015,gallagher-eccv2012,bossard2012apparel,deepfashion,Chen_2015_CVPR}, including ways to jointly recognize attributes~\cite{mixmatch2015,gallagher-eccv2012} or simultaneously detect clothing articles and attributes~\cite{bossard2012apparel}. 
In our work, attributes serve as a starting point for discovering styles (i.e., the ``words" in our topic model), and improvements in attribute prediction from work like the above would also benefit our model.

\vspace*{-0.1in}
\paragraph{Retrieval and matching for clothing}

Given an image of a garment, clothing retrieval methods identify exact or close matches, which supports shopping needs.
Clothing deformations and complex body poses make retrieval challenging, so researchers develop ways to learn the similarity of outfits or garments~\cite{fu2012efficient,street-to-shop2012,getting-the-look2013,runway-realway,lin2015rapid}.
Matching clothing seen ``on the street" to instances stored in catalogs ``at the shop" requires new ideas in domain adaptation~\cite{where-to-buy-iccv2015,huang-cross-domain,getting-the-look2013,street-to-shop2012}.
\CHECK{Existing methods are largely supervised, i.e., provided with street-shop pairs or pairs of outfits judged as similar by human annotators.  In contrast, we develop an unsupervised approach that can leverage ample unlabeled data.  More importantly, whereas retrieval work aims to match the same (or similar) garment(s), we aim to identify style-coherent complete outfits.}

\vspace*{-0.15in}
\paragraph{Models of style, fashionability, compatibility}

Previous work offers a few perspectives on the meaning of visual \emph{style}.  
The Hipster Wars project defines five style categories (\emph{Hipster}, \emph{Goth}, \emph{Preppy}, \emph{Pinup}, \emph{Bohemian})
and recognizes them based on patches on body part keypoints~\cite{hipster}.  Another approach pre-trains a neural network for style using weak meta-data labels~\cite{128floats}.  Compared to the retrieval work above, both (like us) aim to capture a broader notion of style.  However, unlike~\cite{hipster,128floats} we treat styles as discoverable latent factors rather than manually defined categories, which has the advantages discussed in the Introduction.

Whereas style refers to a characterization of whatever it is people wear, \emph{compatibility} refers to how well-coordinated individual garments are~\cite{dyadic,iwata,magic-closet}, and \emph{fashionability} refers to the popularity of clothing items, e.g., as judged by the number of ``like" votes on an image posted online~\cite{fashionability}. Recent work explores forecasting the popularity of styles~\cite{forecast}.

\vspace*{-0.15in}
\paragraph{Topic models}
Topic models originate in text processing.  The well-known topic model Latent Dirichlet Allocation (LDA)~\cite{blei2003latent} and its polylingual extension~\cite{mimno2009polylingual} use multinomial distributions to represent the generation of documents comprised of words. 
Polylingual topic models are applied to Web fashion data in~\cite{vaccaro2016elements} to discover links between textual design elemen meta-data and textual style meta-data, with no computer vision.
Early uses of topic models in vision relied on ``visual words" (quantized image patches) to discover representations for scene recognition~\cite{fei-fei-topics-2005} or perform object category discovery in unlabeled images~\cite{sivic-iccv05}.  More recently, topic models are used to recommend a color-coordinating garment in~\cite{iwata}.  To our knowledge, we are the first to propose discovering visual styles for outfits using topic models.  Deriving topic models on top of semantic visual attributes is also new, and has the added benefit of yielding interpretable latent topics.

\setcounter{figure}{1}
\section{Approach}

After providing background on topic models in \secref{lda_background}, we introduce our visual fashion topic model in \secref{ourtopic}.  Next in \secref{attr_predict} we overview the fine-grained localized attributes used in our model.  Finally, we leverage the learned style-coherent embedding for retrieval, mixing, and summarization tasks in \secref{apps}.


\subsection{Background: Topic models}
\label{sec:lda_background}

We explore unsupervised topic models originating from text analysis to discover visual styles.  In particular, we employ Latent Dirichlet Allocation (LDA)~\cite{blei2003latent}.  LDA is a Bayesian multinomial mixture model that supposes a small number of $K$ latent \emph{topics} account for the distribution of observed words in any given document. It uses the following generative process for a corpus $D$ consisting of $M$ documents each of length $N_{i}$:

\begin{enumerate*}
\item Choose $\boldsymbol{\theta _{i}}\,\sim \,\mathrm {Dir} (\alpha )$, where $i\in \{1,\dots ,M\}$ and $\mathrm {Dir} (\alpha )$ is the Dirichlet distribution for parameter $\alpha$
\item Choose $\boldsymbol{\varphi _{k}}\,\sim \,\mathrm {Dir} (\beta )$, where $k\in \{1,\dots ,K\}$
\item For each word indexed by $i,j$, where $j\in \{1,\dots ,N_{i}\}$, and $i\in \{1,\dots ,M\}$
\begin{enumerate*}
\item Choose a topic $z_{i,j}\,\sim \,\mathrm {Multinomial} (\boldsymbol{\theta _{i}})$
\item Choose a word $x_{i,j}\,\sim \,\mathrm {Multinomial} (\boldsymbol{\varphi _{z_{i,j}}})$
\end{enumerate*}
\end{enumerate*}
Only the word occurrences are observed.

Polylingual LDA~\cite{mimno2009polylingual} extends LDA (we will call it MonoLDA) to process an aligned corpus of documents expressed in multiple languages.  The idea is to recover topics that preserve the ties between translated text.  In particular, translated documents form a tuple, and all documents in a tuple have the same distribution over topics.  Each topic is produced from a set of distributions over words, one distribution per language.

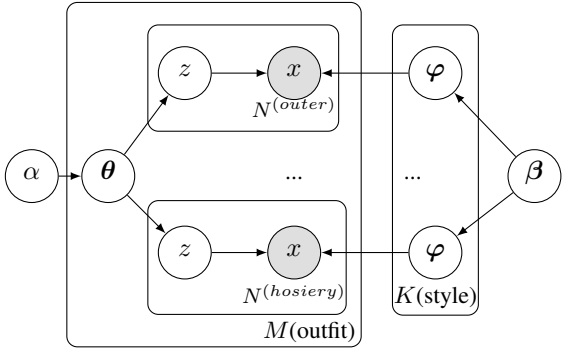
\begin{figure}[t]
  \centering
    \begin{tikzpicture}[>=latex,text height=0.8ex,text depth=0.2ex]
      \small
      \matrix[row sep=0.3cm,column sep=0.3cm] {
        &&\node (z_outer)[latent] {$z$}; &\node (x_outer)[obs] {$x$}; &\node (phi_outer)[latent, xshift=3mm] {$\boldsymbol{\varphi}$};\\ [-3mm]
        &&&\node(caption_outer){\footnotesize{$N^{(outer)}$}};\\
        \node (alpha)[latent] {${\alpha}$}; &\node (theta)[latent] {$\boldsymbol{\theta}$}; &&\node(doc_etc){...}; &\node(topic_etc){...}; &\node (beta)[latent] {$\boldsymbol{\beta}$};\\
        &&\node (z_hosiery)[latent] {$z$}; &\node (x_hosiery)[obs] {$x$}; &\node (phi_hosiery)[latent, xshift=3mm] {$\boldsymbol{\varphi}$};\\ [-2mm]
        &&&\node(caption_hosiery){\footnotesize{$N^{(hosiery)}$}}; &\node(caption_K)[xshift=3mm]{$K$(style)};\\[-2mm]
        &&&\node(caption_M)[xshift=2mm]{$M$(outfit)}; \\
         };
      \path[->]
        (alpha) edge[] (theta)
        (z_outer) edge[] (x_outer)
        (phi_outer) edge[] (x_outer)
        (z_hosiery) edge[] (x_hosiery)
        (phi_hosiery) edge[] (x_hosiery)
        (theta) edge[] (z_outer)
        (theta) edge[] (z_hosiery)
        (beta) edge[] (phi_outer)
        (beta) edge[] (phi_hosiery)
      ;
      \plate {outer} {(z_outer)(x_outer),inner sep=.07cm,xshift=.05cm,yshift=.1cm} {} ;
      \plate {hosiery} {(z_hosiery)(x_hosiery),inner sep=.12cm,xshift=.1cm,yshift=.1cm} {} ;
      \plate {phi} {(phi_outer)(phi_hosiery),inner sep=.1cm,yshift=.07cm} {};
      \plate {document} {(theta)(outer)(hosiery),inner sep=.08cm, yshift=.1cm} {} ;
      \begin{pgfonlayer}{background}
      \end{pgfonlayer}
    \end{tikzpicture}
    \vspace*{-0.15in}
  \caption{Graphical model of the polylingual visual style LDA.}
  \label{fig:graphical_model}
\end{figure}


\subsection{Discovering a style-coherent embedding}
\label{sec:ourtopic}
We propose to learn a style-coherent embedding using a topic model. In our setting, the latent topics will be dis-
covered from unlabeled full-body fashion images, meaning images of people wearing an entire outfit (as opposed to catalog images of individual garments). In this way, we aim to discover the compositions of lower-level visual cues that characterize the main visual themes---€"styles---€"emerging in how people choose to assemble their outfits.

The basic mapping from document topic models to our visual style topic models is as follows: an observed outfit is a ``document", a predicted visual attribute (e.g., \emph{polka dotted}, \emph{flowing}, \emph{wool}) is a ``word", and each style is a discovered ``topic".

\begin{figure}[t]
  \centering
  \footnotesize
  \setlength{\tabcolsep}{0.2em} 
  \ra{1.1}
  \begin{tabular}{@{}ll|ll@{}}
    Mono-topic1 & Mono-topic2 & ~~Poly-topic1 & Poly-topic2\\
    \midrule
    \upp shirt collar & \lowp skirt & \outp deco button & \outp length short\\
    \upp deco button & \lowp skirt short & \outp pattern plain & \outp sleeve long\\
    \upp buttoned & \lowp skirt full & \outp blazer & \outp pullover\\
    \glp deco button & \lowp skirt pleat & \upp buttoned & \upp shirt collar\\
    \upp sleeve long & \lowp skirt high-rise & \upp shirt collar & \upp color white\\
    \glp pattern plaid & \glp pattern plain & \lowp length long & \lowp skirt short\\
    \glp pattern plain & \glp front pullover & \lowp shape straight & \lowp skirt full\\
                       & \glp deco button & \glp deco button & \hosp pattern plain\\
                       &                        & \glp pants & \hosp length short\\
                       &                        & \glp jacket & \glp sweater\\
    {\includegraphics[width=0.12\textwidth]{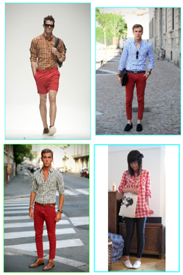}}&
    {\includegraphics[width=0.12\textwidth]{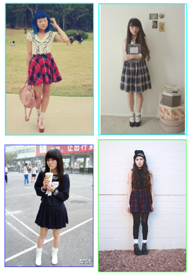}}&
    {\includegraphics[width=0.12\textwidth]{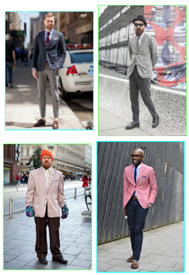}}&
    {\includegraphics[width=0.12\textwidth]{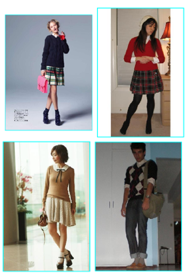}}\\

  \end{tabular}\vspace*{-0.1in}
  \caption{Mono vs.~Polylingual LDA: \upp for \emph{upper body}, \outp for \emph{outer layer}, \lowp for \emph{lower body}, \hosp for \emph{hosiery} and \glp for \emph{global}. MonoLDA (left) learns a topic either for \upp or \lowp, while PolyLDA's styles (right) span the whole body.}
  \label{fig:mono_vs_poly}
\end{figure}

A potential limitation of this mapping, however, is that it treats an outfit as a bag of attributes.  Thus it loses valuable information about the attributes' associated articles of clothing.  For example, learned topics could interchange the appearance of \emph{wool} pants with a \emph{wool} jacket, when the two may in reality signify distinct latent styles.  A partial solution is to specify localized attributes.  For example, we could expand \emph{wool} to \emph{wool pants} and \emph{wool jacket}.  However, this expansion may suffer from allowing LDA to decouple topics across different regions of the body.  For example, in \figref{mono_vs_poly}~(left), MonoLDA dedicates topic 1 to shirt and topic 2 to skirt.

Thus, we consider a polylingual LDA (PolyLDA) model~\cite{mimno2009polylingual}.  In this case, each region of the body is a ``language'', and an outfit is a ``document tuple'' in multiple languages.  As above, latent styles are topics and inferred clothing attributes are words.  The body regions (denoted as $R$) we consider are: \emph{outer layer} (i.e., where a jacket or blazer goes), \emph{upper body} (shirt/blouse/sweater), \emph{lower body} (pants/skirt/shorts), and \emph{hosiery} (tights/leggings).  The polylingual topic model adds a structural constraint that forces body regions to \emph{share} styles, such that we can learn styles consistent across body regions.  
The generative process of PolyLDA is as follows:

\begin{enumerate*}
  \item For each topic $k \in \{1,\dots ,K\}$
  \begin{enumerate*}
    \item For each body part $r \in R$
      \begin{enumerate}[leftmargin=-0.03mm]
        \item Choose attribute distribution $\boldsymbol{\varphi_k^{(r)}} \sim \mathrm{Dir}(\beta)$
      \end{enumerate}
  \end{enumerate*}
  \item For each outfit $i \in \{1,\dots ,M\}$
    \begin{enumerate*}
      \item Choose style distribution $\boldsymbol{\theta_i} \sim \mathrm{Dir}(\alpha)$
      \item For each body part $r \in R$ 
        \begin{enumerate}[leftmargin=-0.03mm]
          \item For each attribute belonging to that body part, indexed by $i, j$, where $j \in \{1,2, \dots ,{N_i}^{(r)}\}$
          \begin{enumerate} [leftmargin=-0.03mm]
            \item Draw a style $z_{ij}^{(r)} \sim \mathrm{Multinomial}(\boldsymbol{\theta_i})$
            \item Draw an attribute $x_{ij}^{(r)} \sim \mathrm{Multinomial}(\boldsymbol{\varphi_{z_{ij}^{(r)}}^{(r)}})$
          \end{enumerate}
        \end{enumerate}
    \end{enumerate*}
\end{enumerate*}

\figref{graphical_model} shows the associated graphical model.
In contrast to MonoLDA, PolyLDA captures the interaction of garment regions, such that each style specifies a full-body trend (\figref{mono_vs_poly}, right).

By applying PolyLDA to a database of unlabeled outfit images, we obtain a set of discovered styles (see \figref{lda_top_images} in our experiment section for examples) with which to encode novel images.
Each topic $k$ has its attribute probability $\boldsymbol{\varphi_k^{(r)}}$ depending on body region $r$.  Given an outfit $d$, we represent it in a style-coherent embedding by its topic proportions:
\vspace*{-0.09in}
\begin{equation}
\boldsymbol{\theta}_d =[\theta_{d1},\dots,\theta_{dK}],
\vspace*{-0.09in}\end{equation}
where $\theta_{dk}\geq 0$, $\Sigma_k \theta_{dk} = 1$.
The resulting embedding accounts for the fact that a \emph{composition} of style elements  defines a look~\cite{barthes2013language}.

We stress that our style-coherent embedding is fully unsupervised. Our method discovers styles from unlabeled images, as opposed to learning a style embedding with supervision. For example, one could gather pairs of fashion images and ask human annotators to label them as similar or dissimilar in style (or use noisy tags as labels [33]), then learn an embedding that keeps similar pairs close. Or, in the spirit of [19], one could train classifiers to target a pre-defined set of style categories.   While the attribute models are trained on a disjoint pool of attribute labeled images, our style model runs on \emph{predicted} attributes; annotators do not touch the images on which we perform discovery.  Our unsupervised strategy saves manual effort. More importantly, it also addresses challenges specific to visual styles---namely, their ever-evolving nature, the difficulty in enumerating them with words, and their soft boundaries.


\subsection{Fine-grained localized fashion attributes}
\label{sec:attr_predict}
\begin{figure}[t]
   \center
   \includegraphics[width=0.95\linewidth]{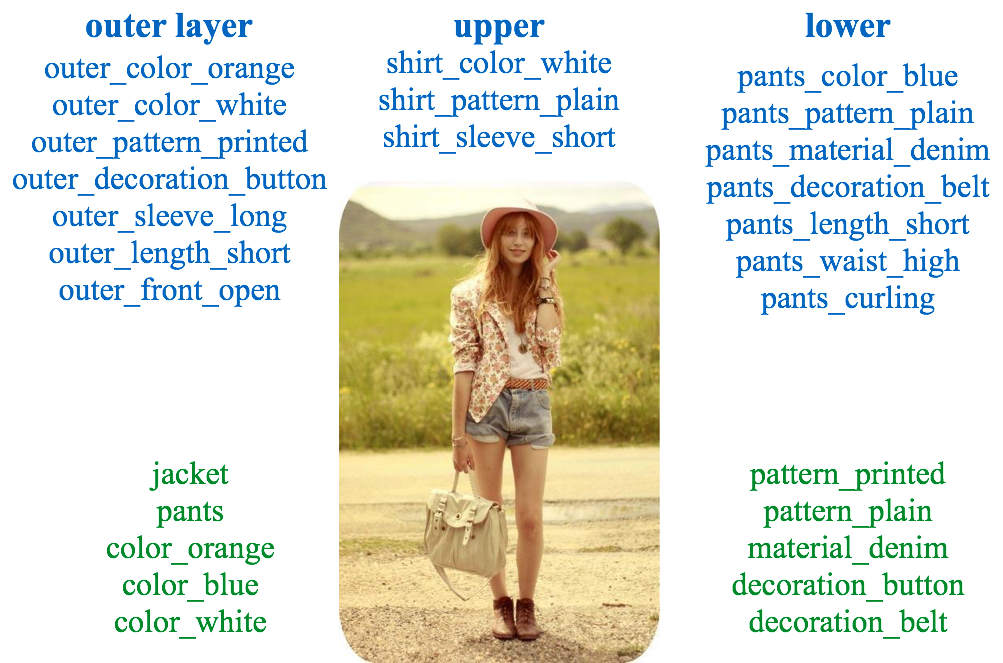}
   \caption{Attributes present for an outfit in the \emph{\textcolor{skyblue2}{localized}} vocabulary (top) or \emph{\textcolor{chameleon3}{global}} vocabulary (bottom).}
   \label{fig:attr_ex}
\end{figure}
We next discuss our approach to infer attributes in full-body fashion images. We consider both global and localized attributes. Global attributes indicate the presence of a property somewhere on the body (e.g., \emph{floral}), whereas a localized attribute links it specifically to a body region (e.g., \emph{floral-shirt} and \emph{floral-skirt} are distinct words).  \figref{attr_ex} shows an example image and the attributes from either vocabulary that are present, as well as the body region association for the localized ones. 
\vspace*{-0.15in}
\paragraph{Vocabulary and data collection}

As input to our style discovery model, we need a rich attribute vocabulary that is both localized and fine-grained.
In existing fashion datasets, the attributes lack one or both of these aspects~\cite{bossard2012apparel,gallagher-eccv2012,di2013style,mixmatch2015} or are not publicly available~\cite{street-to-shop2012}.  Thus, we curate a new dataset for attribute training.

For the vocabulary, we build on the 53 attributes enumerated in~\cite{street-to-shop2012}.  First we remove those too subtle for most annotators to discern (\emph{chiffon}; \emph{jewel} collar).  Then we add missing but frequently appearing attributes (e.g.,
\emph{pink}; \emph{polka-dot}). Finally, we expand the set so that \emph{color}, \emph{material}, and \emph{pattern} are localized to each body region.  This yields 195 total attributes (see Supp for details).

To gather images, we use keyword search with the attribute name on Google, then manually prune those where the attribute is absent.
This yields 70 to 600 positive training images per attribute. We also gather 2000 random street images from chictopia.com (manually pruned for false-negatives) to serve as negative examples.  In total, the new dataset has 18,878 images.\footnote{\url{vision.cs.utexas.edu/projects/StyleEmbedding/}}

\vspace*{-0.1in}
\paragraph{Training with multilabel outfits}

The clothing outfit images are multilabel in terms of their attributes.  To circumvent the expense of labeling all 19K images for all 195 attributes, and to deal appropriately with highly localized attributes, we develop a piecewise training procedure.  First we group the attributes into six types: \emph{pattern}, \emph{material}, \emph{shape}, \emph{collar type}, \emph{clothing article}, and \emph{color}.  The types have 105, 15, 20, 8, 27, 13 attributes, respectively.\footnote{Attributes within \emph{pattern} and \emph{collar} types are mutually exclusive, thus their multilabeling can be done efficiently.  We obtain complete 20-label and 15-label multilabeling for types \emph{material} and \emph{shapes}.}
Then we train separate convolutional neural networks (CNN) per type.  This allows us to directly use the positive examples for each attribute, and all others from other keywords as negatives, while still yielding predictions at test time that are multilabel.  \CHECK{We find it is also important during training to have negatives with none of the named attributes present, since such outfits are rather common.}

Our attribute learning framework accounts for the challenge that many attributes occupy a small portion of a fashion image.  First, we detect people using faster-RCNN~\cite{NIPS2015_5638}, and extract a crop from the detected person bounding box according to whether the image is a training image for an upper or lower body attribute. We give the network both the whole-body person box and the crop as two instances with the same attribute label, allowing it to leverage any useful cues. We fine-tune our attribute prediction networks (one per type for the first four types) on ResNet-50~\cite{he2016deep} pre-trained with ImageNet~\cite{imagenet}.

\begin{figure}[t]
\centering
   \includegraphics[width=0.9\linewidth]{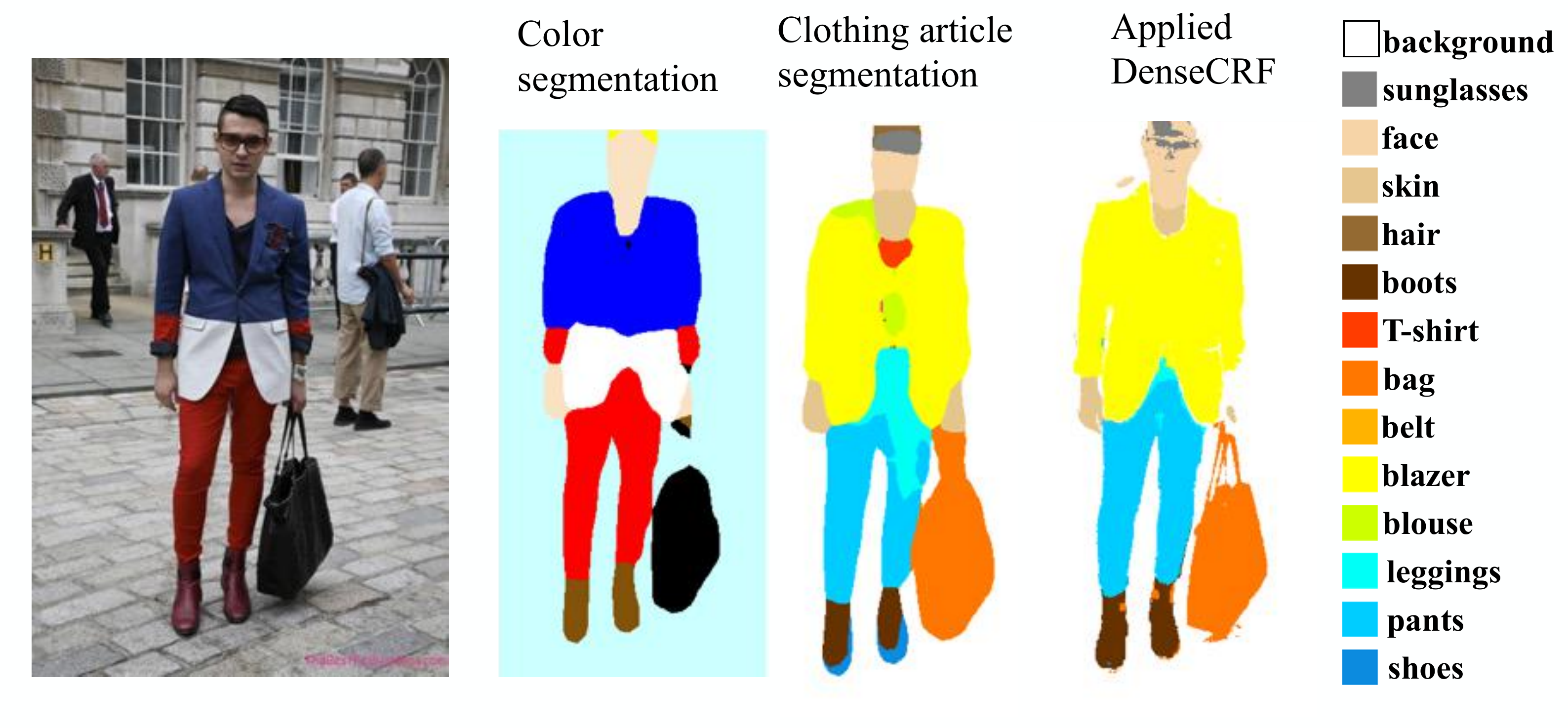}
   \vspace*{-0.15in}
   \caption{
   We intersect the article and color labels to generate the final \emph{blazer-color-blue}, \emph{pants-color-red} prediction.}
   \label{fig:segmentation_ex}
\vspace*{-0.1in}
\end{figure}

For both the \emph{clothing article} and \emph{color} types, we train a segmentation network.  For both, we fine-tune DeepLab's~\cite{CP2015Semantic} repurposed VGG-16 network and apply DenseCRF~\cite{KrahenbuhlK11}. The networks target 27 pixel-wise clothing article labels and 13 pixel-wise color labels from the Fashionista data~\cite{paperdoll-iccv2013}, respectively.  At test time, we i) record the detected clothing article names, and ii) intersect the color and clothing semantic segmentations to produce article-specific color attributes, e.g., \emph{shirt-color-blue} (Fig.~\ref{fig:segmentation_ex}).

\begin{figure}[t]
  \footnotesize
    \centering
    \setlength{\tabcolsep}{0.4em}
    \begin{tabular}{@{}llllll@{}}
      &  Upper &  Lower  &  Hosiery \quad \quad & \multicolumn{2}{c}{Global} \\
      \midrule
     
      \multirow{4}{*}{\includegraphics[width=1.4cm]{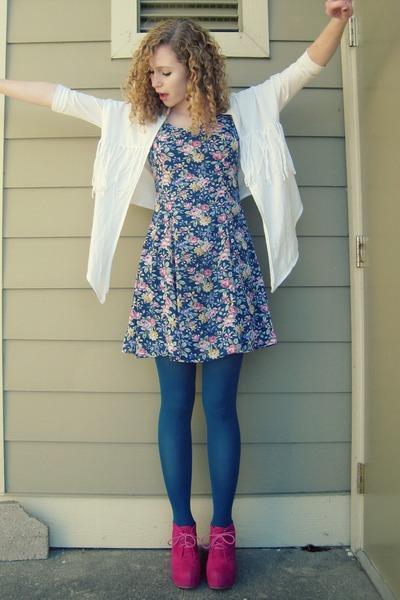}}
      & blue & short dr. & blue & floral & blue\\
      & purple & loose dr. && translucent & purple\\
      & white & flat dr. & & dress & white\\
      & & white dr. & & shoe & beige\\
      & & blue dr. & & stocking & cardigan\\
      & & purple dr. & & red \vspace{10pt}\\
      
     \multirow{4}{*}{\includegraphics[width=1.4cm]{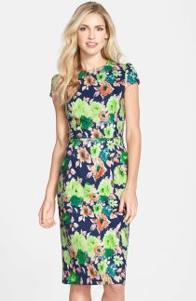}}
     & floral & floral dr. && floral\\
     & blue & tight dr. && dress\\
     & green & flat dr. && blue\\
     & beige & blue dr. && green\\
     & & green dr.&& beige\\
     & & beige dr.\\
     &&&&&\\
    \end{tabular}\vspace*{-0.15in}
  \caption{Example of predicted attributes on a HipsterWars~\cite{hipster} (top) image and a DeepFashion~\cite{deepfashion} (bottom) image (dr.=dress)}
  \label{fig:predicted_attr_ex}
\end{figure}

\begin{figure*}[t]
   \begin{center}
   \includegraphics[width=.95\linewidth]{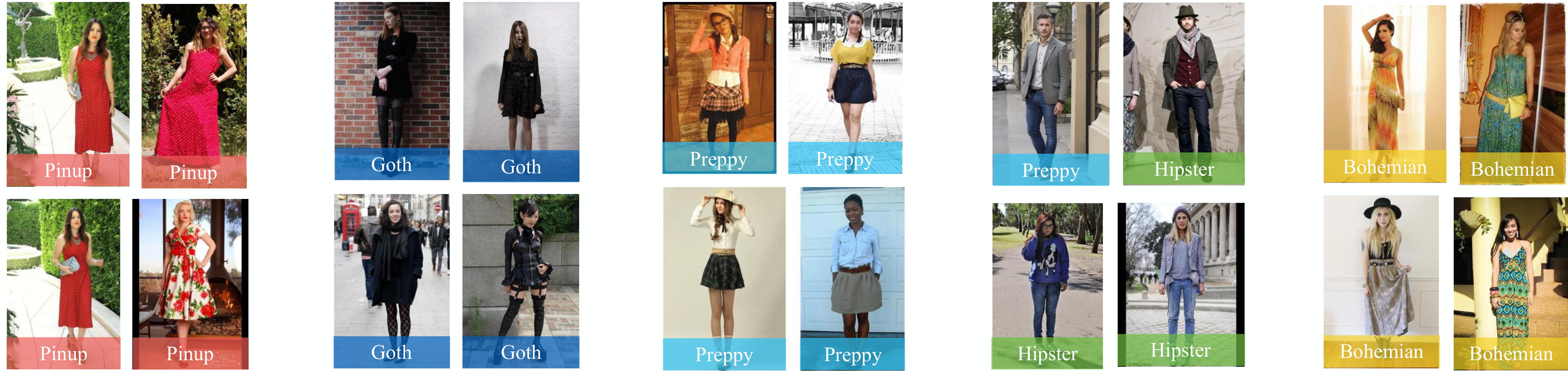}
   \caption{Top images for the five \textbf{discovered style topics} with PolyLDA. Labels indicate human-assigned styles from HipsterWars~\cite{hipster}, which are \emph{not} seen by our algorithm. Our approach successfully discovers the five human-perceived styles.  Please see Supp for more examples and baseline cluster results.} 
   \label{fig:lda_top_images}
   \end{center}\vspace*{-0.1in}
\end{figure*}

The resulting attribute classifiers offer a fairly reliable basis for style discovery.  For the validation split of our  19K image dataset, they attain $90\%$ average precision.  \figref{predicted_attr_ex} shows attribute predictions on novel test images.

\subsection{Using the style-coherent embedding}
\label{sec:apps}

Our method produces a style-coherent representation for fashion images.
Our experiments consider a few tasks leveraging this representation:

\vspace*{-0.1in}
\paragraph{Retrieval of style-related images}

For retrieval, the system is given a query image and must return database images that illustrate similar style. 
Here we simply use our learned embedding to retrieve images close to a query image, i.e., nearest neighbors in the space of $\boldsymbol{\theta}_q$ for query image $q$.

Our embedding retrieves images that maintain style coherence with the query.  While conventional embeddings (e.g., CNNs) can return the examples \emph{closest in appearance}, our embedding can return those that are \emph{close in style}.  Whereas the former is preferable when doing street-to-shop visual matching~\cite{where-to-buy-iccv2015,huang-cross-domain,getting-the-look2013,street-to-shop2012}, the latter is preferable for many browsing scenarios, e.g., to view recommendations related to past purchases.

\vspace*{-0.1in}
\paragraph{Mixing and ``traversing between" styles}

A new task supported by our approach is to \emph{mix} fashion styles.  In this scenario, a user identifies $T$ styles of interest ${\mathbf S}:=\{S_1,\dots,S_T\}$, $S_t \in \{1,\dots,K\}$, and queries for images that exhibit a blend of those styles.  For example, the user could manually select styles of interest by viewing images associated with each discovered style, or the styles for mixing could be automatically discovered based on the dominant styles in his photo album or shopping history.  We measure the relevance of an image $I_i$ as:
\begin{equation}
\textrm{MixRelevance}(\boldsymbol{\theta}_i, {\mathbf S}) = \min_{t \in {\mathbf S}} \theta_{it},
\end{equation}
where $\boldsymbol{\theta}_i$ is the style embedding for $I_i$.  \CHECK{The min assures that an image is only as relevant to the requested style mix as it is close to its most distant style.}  Similarly, we can offer new browsing capabilities by depicting a gradual transition from one style to another (see \figref{mix_portion}).

\vspace*{-0.1in}
\paragraph{Summarizing styles}

A third application uses our style-coherent embedding to summarize the styles in an image collection.  Given images $\{I_1,\dots,I_N\}$, we calculate the relative influence of each style $k$ as $\textrm{Influence}(S_k) = \sum_{i=1}^N \boldsymbol{\theta}_{ik}$.  With these frequencies we can visualize the collection compactly by sampling images dominant for each influential style (\figref{summaries} in our experiment section).

\begin{table}
   \centering
   \begin{tabular}{@{}lcc@{}}
      & {Avg. max AP} & {NMI}  \\
      \midrule
      \quad MonoLDA & 0.48 & 0.30 \\
      \quad PolyLDA & \bf0.53\vspace{5pt} & \bf{0.31}
   \end{tabular}
   \vspace{-2mm}
   \caption{Mono vs.~poly LDA discovery accuracy judged against the manual style labels of HipsterWars, using GT attributes.}
   \label{tab:topic_model_comparison}
\end{table}

\section{Experiments}\label{sec:results}

\cc{We first show that our discovered topics align with human-perceived styles (\secref{exp_topic_effectiveness}). Then, we apply the embedding for retrieval (\secref{exp_style_retrieval}), mixing  (\secref{exp_mixing-results}), and summarizing styles (\secref{exp_summary-results}).}

\vspace*{-0.1in}
\paragraph{Datasets}  We use two datasets: (i) HipsterWars~\cite{hipster}, which has 1,893 images, each labeled by one of 5 style labels: Hipster, Preppy, Goth, Pinup, Bohemian; (ii) DeepFashion~\cite{deepfashion}, from which we take all 108,145 images that have at least one of the 230 style labels \emph{and} a fully-visible person.  Because the 230 style labels in DeepFashion are noisy labels, we collapse them into 42 higher level styles by affinity propagation~\cite{affinitypropagation} using cosine similarity to measure co-occurrence of styles in an outfit (see Supp).  We use the attribute networks trained with our new 19K image dataset (Section~\ref{sec:attr_predict}) to predict the attributes in the HipsterWars and DeepFashion images.

\vspace*{-0.1in}
\paragraph{Baselines} We compare with four baseline: (i) StyleNet~\cite{128floats}: a state-of-the-art feature for clothing that fine-tunes a CNN using 123 metadata labels (e.g., red-sweater) on images from the Fashion 144K dataset~\cite{fashionability}, (ii) vanilla ResNet-50~\cite{he2016deep}: the last layer of a state-of-the-art CNN pretrained for ImageNet, (iii) Attr-ResNet: ResNet-50 fine-tuned to classify the same 148 attributes\footnote{The attributes in types: \emph{pattern}, \emph{material}, \emph{shape}, \emph{collar type}; we did not include \emph{clothing article} and \emph{color} because they are predicted differently, from a segmentation network.} used by our method with the same training data, and (iv) Attributes: indicator vectors using the same attributes as our model. 

\subsection{Consistency with human labeled styles}
\label{sec:exp_topic_effectiveness}

First we analyze how well our discovered styles align with human perception, as captured by the datasets' style labels (never seen by our approach).  We use two metrics: (i) Normalized Mutual Information (NMI)\cc{, which captures the overall alignment of topics with ground truth (GT) styles}; and (ii) averaged maximal average precision (AP) per style, which uses each topic's probability as a relevance score for a style to sort all images, then records the AP per topic per style. The best (max) AP a style has in all topics is that style's final score.  We average the max AP scores of all styles (5 in HipsterWars and 200+ in DeepFashion) to get ``avg maximal AP''. To extract clusters for the baseline representations we use $K$-means clustering.  \CHECK{We also tried GMM and AP-clustering and found the clustering algorithm itself has negligible impact on their results.}  For all methods, we set the number of clusters/topics to be the number of style labels in the respective dataset.

First we examine the impact of our polylingual model.  \tabref{topic_model_comparison} shows the results for both LDA variants on HipsterWars.  Here we use ground truth attributes, in order to evaluate the LDA models independent of attribute prediction quality. 
We see that the polylingual model has an advantage, and thus adopt it as our model for all experiments.

Next we quantify discovery accuracy for our approach against the baselines.
\tabref{baseline_comparison} shows the results on both datasets.  For the attribute-indicator baseline and our approach, we show results with predicted and ground truth attributes in order to separate the success of discovery from the success of attribute prediction.\footnote{DeepFashion has GT for only global attributes.}  
Overall, PolyLDA is \weilin{the} strongest.  Both PolyLDA and Attributes perform better with perfect attributes, reinforcing that attribute precision is an important ongoing research challenge~\cite{mixmatch2015,gallagher-eccv2012,bossard2012apparel}.  
However, even with predicted attributes, we outperform all baselines on \weilin{both} datasets for both metrics.  
Despite having been pretrained to capture noisy fashion labels, the StyleNet CNN~\cite{128floats} does not discover the human-perceived styles as well, though it does soundly outperform the vanilla ResNet baseline. \weilin{Attr-ResNet falls in between StyleNet and vanilla ResNet, as expected.}
The absolute accuracy on DeepFashion is much lower for all methods, a function of its larger size and more varied and noisy style labels.  Whereas the Hipster style labels are manually curated through a rigorous crowdsourcing procedure~\cite{hipster}, the DeepFashion style labels are gleaned from text meta-data~\cite{deepfashion}.

\figref{lda_top_images} shows the most central images for our discovered styles on HipsterWars (see Supp for DeepFashion).  The qualitative examples reinforce the quantitative result above.  Our model discovers the human-perceived stylesbetter than the CNN and attributes clusters (see Supp).  Our style-coherent embedding better tolerates superficial visual differences in intra-style images. 


\subsection{Style-coherent retrieval}\label{sec:exp_style_retrieval}

Having shown that the discovered styles are meaningful, next we evaluate the style embedding for retrieval.  
In this task, a user queries by example for images related by style, e.g., for recommendation or catalog browsing relative to some currently viewed item (query).  Recall, this is distinct from instance retrieval for near-duplicates.  Thus, we evaluate performance simultaneously by \emph{style coherence} and \emph{diversity} or \emph{novelty}.  Diversity refers to the retrieved images' mutual visual dissimilarity with each other, and novelty refers to their collective dissimilarity to the query.  The goal is to obtain retrieval results that maintain style coherence while avoiding redundancy.  For style coherence, we evaluate NDCG~\cite{ndcg} against ground truth style labels.  For diversity/novelty, we learn a metric to mimic human-perceived dissimilarity, following~\cite{whittlesearch}.  In particular, we collect 350 triplets labeled by 5 human annotators and learn a ranking function~\cite{ranksvm} on top of the attribute and CNN descriptors that respects human-given judgments.

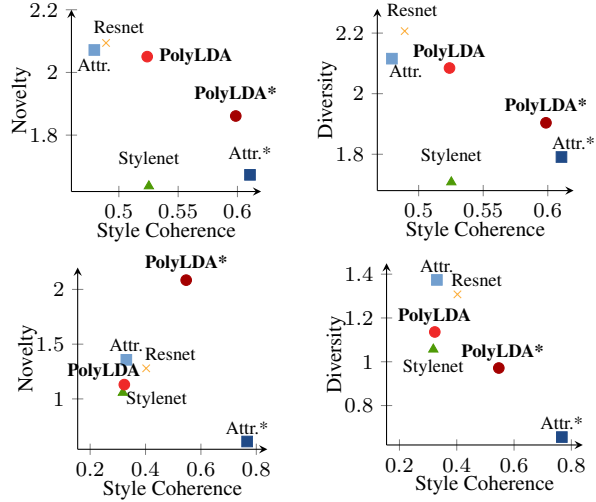
\begin{figure}
\footnotesize
\hspace*{-0.2in}
\begin{tabular}{c}
   \begin{subfigure}{0.235\textwidth}
   \center
   \begin{tikzpicture}
       \begin{axis}[xlabel=Style Coherence,
                 ylabel=Novelty,
                 xlabel shift = -4pt,
                 ylabel shift = -6pt,
                 axis lines=left,
                 xmin=0.46,
                 ymin=1.62,
                 xmax=0.62,
                 ymax=2.2,
                  width=1.0\textwidth,
                 height=4cm
                ]
           \addplot[
                   visualization depends on={value \thisrow{name}\as\myvalue},
               scatter/classes={
                   b_data={mark=square*,skyblue1,text mark=\myvalue},
                   c_data={mark=triangle*,chameleon3,text mark=\myvalue},
                   d_data={mark=x,orange1,text mark=\myvalue},
                   f_data={mark=square*,skyblue3,text mark=\myvalue},
                   b_name={mark=x,white,text mark=\myvalue},
                   c_name={mark=x,white,text mark=\myvalue},
                   d_name={mark=x,white,text mark=\myvalue},                   
                   f_name={mark=x,white,text mark=\myvalue}
                   },
                   scatter, only marks,
                   scatter src=explicit symbolic,
                   nodes near coords*={\myvalue},font=\scriptsize,]
            table[x=x,y=y,meta=label]{hipster_coher_novelty_baseline.txt};
             \addplot[
                   visualization depends on={value \thisrow{name}\as\myvalue},
               scatter/classes={
                   a_data={mark=*,scarletred1,text mark=\myvalue},
                   e_data={mark=*,scarletred3,text mark=\myvalue},
                   a_name={mark=*,white,text mark=\myvalue},
                   e_name={mark=*,white,text mark=\myvalue}
                   },
                   scatter, only marks,
                  scatter src=explicit symbolic,
                  nodes near coords*={\textbf{\myvalue}},font=\scriptsize,]
            table[x=x,y=y,meta=label]{hipster_coher_novelty_ours.txt};
       \end{axis}
   \end{tikzpicture}
   \end{subfigure}\hfill
   \begin{subfigure}{0.24\textwidth}
   \center
   \begin{tikzpicture}
       \begin{axis}[xlabel=Style Coherence,
                 ylabel=Diversity,
                 xlabel shift = -4pt,
                 ylabel shift = -6pt,
                 axis lines=left,
                 xmin=0.468,
                 ymin=1.68,
                 xmax=0.62,
                 ymax=2.28,
                  width=1.0\textwidth,
                 height=4cm
                ]
           \addplot[
                   visualization depends on={value \thisrow{name}\as\myvalue},
               scatter/classes={
                   b_data={mark=square*,skyblue1,text mark=\myvalue},
                   c_data={mark=triangle*,chameleon3,text mark=\myvalue},
                   d_data={mark=x,orange1,text mark=\myvalue},
                   f_data={mark=square*,skyblue3,text mark=\myvalue},
                   b_name={mark=x,white,text mark=\myvalue},
                   c_name={mark=x,white,text mark=\myvalue},
                   d_name={mark=x,white,text mark=\myvalue},                   
                   f_name={mark=x,white,text mark=\myvalue}
                   },
                   scatter, only marks,
                   scatter src=explicit symbolic,
                   nodes near coords*={\myvalue},font=\scriptsize,]
            table[x=x,y=y,meta=label]{hipster_coher_diversity_baseline.txt};
            \addplot[
                   visualization depends on={value \thisrow{name}\as\myvalue},
               scatter/classes={
                   a_name={mark=x,white,text mark=\myvalue},
                   e_name={mark=x,white,text mark=\myvalue},
                   a_data={mark=*,scarletred1,text mark=\myvalue},
                   e_data={mark=*,scarletred3,text mark=\myvalue}
                   },
                   scatter, only marks,
                   scatter src=explicit symbolic,
                   nodes near coords*={\textbf{\myvalue}},font=\scriptsize,]
            table[x=x,y=y,meta=label]{hipster_coher_diversity_ours.txt};
       \end{axis}
   \end{tikzpicture}
   \end{subfigure}\\
   \begin{subfigure}{0.235\textwidth}
   \center
   \begin{tikzpicture}
       \begin{axis}[xlabel=Style Coherence,
                ylabel=Novelty,
                xlabel shift = -4pt,
                ylabel shift = -6pt,
                 axis lines=left,
                 xmin=0.155,
                 ymin=0.54,
                 xmax=0.84,
                 ymax=2.2,
                  width=1.0\textwidth,
                 height=4cm
                ]
            \addplot[
                   visualization depends on={value \thisrow{name}\as\myvalue},
               scatter/classes={
                   b_data={mark=square*,skyblue3,text mark=\myvalue},
                   c_data={mark=triangle*,chameleon3,text mark=\myvalue},
                   d_data={mark=x,orange1,text mark=\myvalue},
                   f_data={mark=square*,skyblue1,text mark=\myvalue},
                   b_name={mark=*,white,text mark=\myvalue},
                   c_name={mark=*,white,text mark=\myvalue},
                   d_name={mark=*,white,text mark=\myvalue},
                   f_name={mark=*,white,text mark=\myvalue}
                   },
                   scatter, only marks,
                   scatter src=explicit symbolic,
                   nodes near coords*={\myvalue},font=\scriptsize,]
            table[x=x,y=y,meta=label]{large_coher_novelty_baseline.txt};
            \addplot[
                   visualization depends on={value \thisrow{name}\as\myvalue},
               scatter/classes={
                   a_name={mark=*,white,text mark=\myvalue},
                   e_data={mark=*,scarletred1,text mark=\myvalue},
                   e_name={mark=*,white,text mark=\myvalue},
                   a_data={mark=*,scarletred3,text mark=\myvalue}
                   },
                   scatter, only marks,
                   scatter src=explicit symbolic,
                   nodes near coords*={\textbf{\myvalue}},font=\scriptsize,]
            table[x=x,y=y,meta=label]{large_coher_novelty_ours_debugged.txt};
       \end{axis}
   \end{tikzpicture}
   \end{subfigure}\hfill
   \begin{subfigure}{0.24\textwidth}
   \center
   \begin{tikzpicture}
       \begin{axis}[xlabel=Style Coherence,
                 ylabel=Diversity,
                 xlabel shift = -4pt,
                 ylabel shift = -6pt,
                 axis lines=left,
                 xmin=0.155,
                 ymin=0.62,
                 xmax=0.84,
                 ymax=1.45,
                  width=1.0\textwidth,
                 height=4cm
                ]
            \addplot[
                   visualization depends on={value \thisrow{name}\as\myvalue},
               scatter/classes={
                   b_name={mark=*,white,text mark=\myvalue},
                   c_name={mark=*,white,text mark=\myvalue},
                   d_name={mark=*,white,text mark=\myvalue},
                   f_name={mark=*,white,text mark=\myvalue},
                   b_data={mark=square*,skyblue3,text mark=\myvalue},
                   c_data={mark=triangle*,chameleon3,text mark=\myvalue},
                   d_data={mark=x,orange1,text mark=\myvalue},
                   f_data={mark=square*,skyblue1,text mark=\myvalue}
                   },
                   scatter, only marks,
                   scatter src=explicit symbolic,
                   nodes near coords*={\myvalue},font=\scriptsize,]
            table[x=x,y=y,meta=label]{large_coher_diversity_baseline.txt};
            \addplot[
                   visualization depends on={value \thisrow{name}\as\myvalue},
               scatter/classes={
                   a_name={mark=*,white,text mark=\myvalue},
                   e_data={mark=*,scarletred1,text mark=\myvalue},
                   e_name={mark=*,white,text mark=\myvalue},
                   a_data={mark=*,scarletred3,text mark=\myvalue}
                   },
                   scatter, only marks,
                   scatter src=explicit symbolic,
                   nodes near coords*={\textbf{\myvalue}},font=\scriptsize,]
            table[x=x,y=y,meta=label]{large_coher_diversity_ours_debugged.txt};
       \end{axis}
   \end{tikzpicture}
   \end{subfigure}
   \end{tabular}
   \vspace*{-0.1in}
\caption{Style retrieval on HipsterWars (top) and DeepFashion (bottom). * denotes use of GT attributes. 
The ideal method would sit in the top right corner of the plots.  Our embedding offers a good trade-off in style coherency and diversity/novelty.}
\label{fig:retrieval_both}
\end{figure}

\figref{retrieval_both} shows the results for both datasets.  
For HipsterWars (top), we treat each image as a query in turn, and for DeepFashion (bottom) we sample 2,000 of the 108,145 images as queries.
Our model offers a good combination of coherency and diversity/novelty.  
On HipsterWars, it maintains diversity/novelty while maintaining a similar or better level of coherence as the baselines.
As before, predicted attributes diminish coherence, yet the topic model coherence appears to degrade more gracefully.

\begin{table}\hspace*{-0.12in}
   \footnotesize
   \centering
   \begin{tabular}{@{}lcc|cc@{}}
      &\multicolumn{2}{c}{HipsterWars}&\multicolumn{2}{c}{DeepFashion}\\
      \cmidrule{2-3}\cmidrule{4-5}
      & {Avg AP} & {NMI} & {Avg AP} & {NMI} \\
      \midrule
      \quad StyleNet~\cite{128floats} & 0.39 & 0.20 & 0.0501 & 0.0011\\
      \quad ResNet~\cite{he2016deep} & 0.30 & 0.16 & 0.0524 & 0.0004\\
      \quad Attr-ResNet & 0.35 & 0.18 & 0.0615 & 0.0002\\
      \quad Attributes & 0.28~/~0.32 & 0.19~/~0.28 & 0.0560~/~0.1294 & 0.0017~/~0.0082\\
      \quad PolyLDA & 0.50~/~\bf0.53 & 0.21~/~\bf0.31 & 0.0647~/~\bf0.1762 & 0.0116~/~\bf0.0227
   \end{tabular}
   \vspace{-2mm}
   \caption{Discovery accuracy for both datasets.  Attributes and PolyLDA show result if using either predicted attributes (first) or ground truth attributes (second).}
   \label{tab:baseline_comparison}
\end{table}


\subsection{Mixing styles}\label{sec:exp_mixing-results}

Next we consider mixing styles.  Since evaluation of mixing requires images labeled for multiple human-perceived styles as well as instances exhibiting exclusively one style, we collect a ground truthed test set of 177 Web images using the HipsterWars style names (see Supp).  While our mixing approach (Sec~\ref{sec:apps}) can blend arbitrary selected styles, for sake of evaluation we focus on blending pairs of GT-labeled styles, then score the AP against the ground truth, i.e., images exhibiting both the initial selected styles.  For the baselines, we use their clusters analogously to our topics, creating $K$-dim embeddings that record the distance of the image to each cluster's centroid.  We use $K=25$ topics/clusters; $K \in (15,30)$ gives similar results.

\tabref{mix_style} shows the results.  On the whole, our approach does better than the baselines, \CHECK{and in most cases this is true even using predicted attributes.}  This result highlights the power of the topic model over the raw attributes, \CHECK{which are too low-level for adequate mixing.} 

\begin{figure*}[t] 
   \vspace{-7mm}
   \begin{center}
 
        \includegraphics[width=0.9\linewidth]{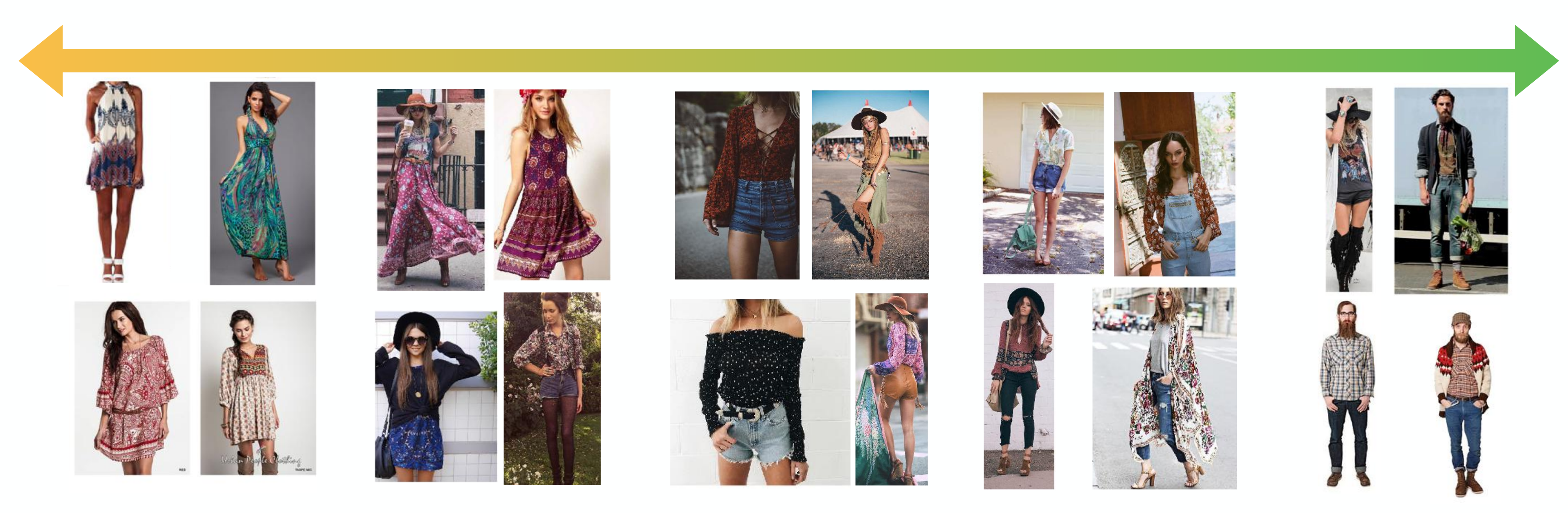}

   \end{center}\vspace*{-0.3in}
   \caption{\small{Visualization generated by mixing our style topics, gradually traversing from one style (Bohemian) to another (Hipster).}}
   \label{fig:mix_portion}
\end{figure*}

\begin{figure} 
   \begin{center}
      \begin{subfigure}[t]{.44\textwidth} 
            \includegraphics[width=1.0\linewidth]{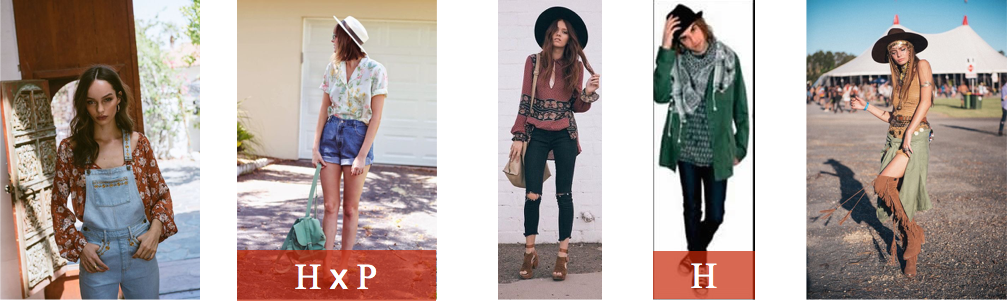}
            \caption{\small{Hipster$\times$Bohemian}}
      \end{subfigure}
      \begin{subfigure}[t]{.44\textwidth} 
            \includegraphics[width=1.0\linewidth]{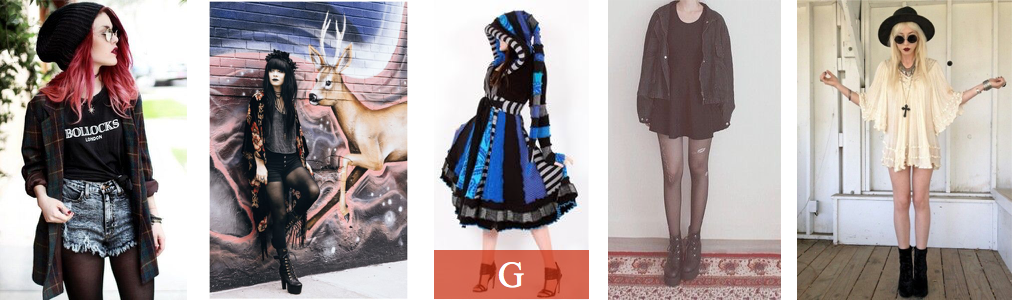}
            \caption{\small{Hipster$\times$Goth}}
      \end{subfigure}
   \end{center}
   \vspace*{-0.2in}
   \caption{Top retrievals for two mixes. Incorrect images are labeled with red actual labels:  \emph{\textbf{H}ipster}, \emph{\textbf{G}oth}, \emph{\textbf{P}reppy}.}
   \label{fig:best_mixture}
\end{figure}

\figref{best_mixture} shows example images predicted as strong exemplars for two style blends. \figref{mix_portion} shows an example gradually mixing from a source style to a target style.

\subsection{Style summaries}\label{sec:exp_summary-results}

\begin{figure} 
   \begin{center}
   \begin{tabular}{c}
      \begin{subfigure}[t]{0.21\textwidth} 
            \includegraphics[width=1\linewidth]{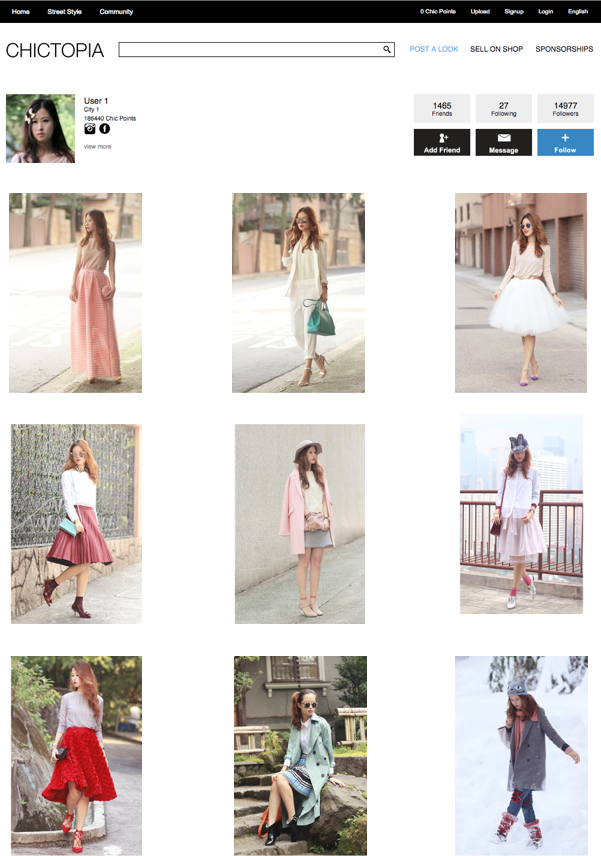}
      \end{subfigure}
      \begin{subfigure}[t]{0.26\textwidth} 
            \includegraphics[width=1\linewidth]{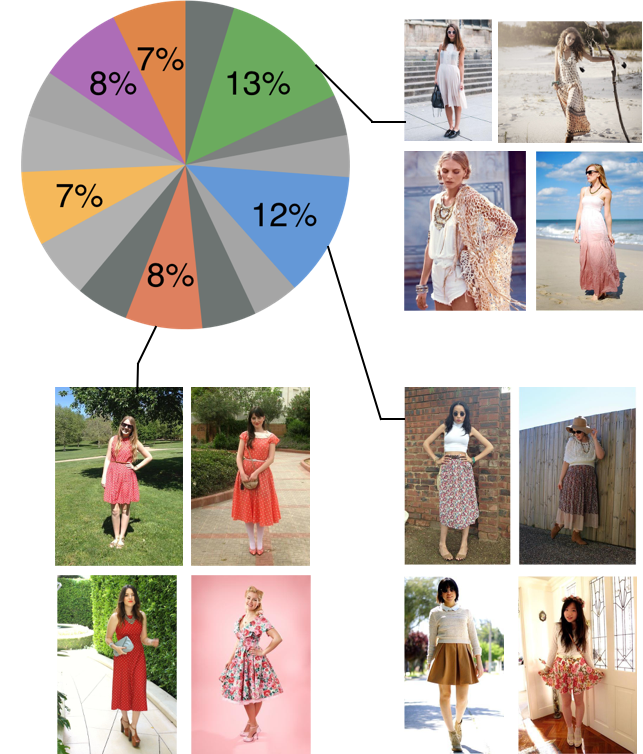}
      \end{subfigure}
      \\\vspace*{0.15in}
       \begin{subfigure}[t]{0.21\textwidth} 
            \includegraphics[width=1\linewidth]{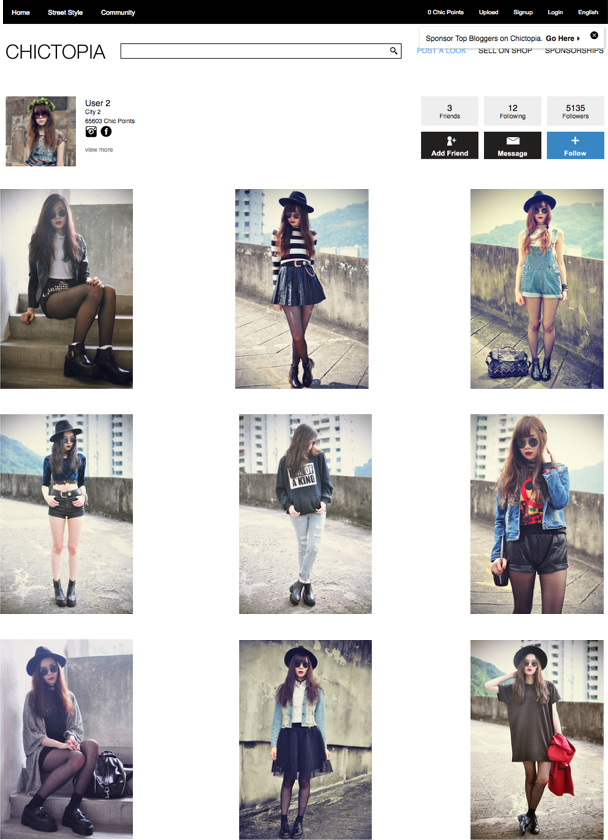}
      \end{subfigure}
      \begin{subfigure}[t]{0.26\textwidth} 
            \includegraphics[width=1\linewidth]{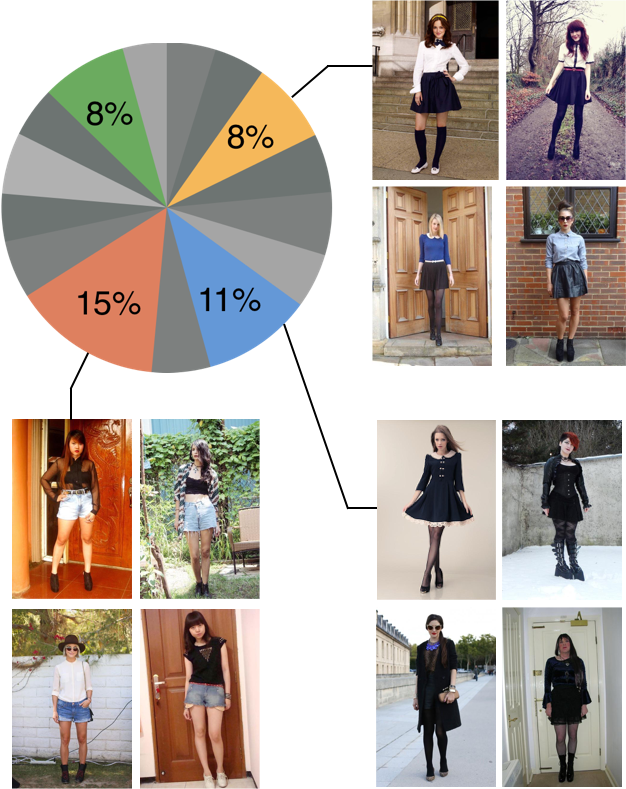}
      \end{subfigure}
      \end{tabular}
   \end{center}
   \vspace{-11mm}
   \caption{Style summarization for two users.  Left is the user's album, right is the visual style summary breaking down the main trends discovered in the album.}
   \label{fig:summaries}
\end{figure}

Finally, we demonstrate the power of our model to organize galleries of outfits.   As proof of concept, we select two users from chictopia.com, and download 200 photos from each of their albums. \figref{summaries} shows the results.   We show snapshots from their albums along with summary piecharts computed by our approach to highlight the dominant styles. Gray pie slices indicate insignificant styles for a user.  Our summaries convey the user's tastes in a glance. In contrast, the status quo would entail manually paging through all 200 photos in the album in an arbitrary order.

\begin{table}\hspace*{-0.25in}
   \scriptsize
   \setlength{\tabcolsep}{0.4em} 
   \begin{tabular}{lllllll} 
      & Preppy$\times$& Hipster$\times$ & Preppy$\times$ &  Goth$\times$ & Bohemian$\times$\\
      & Goth & Goth & Hipster & Bohemian & Hipster\\
      \cmidrule(l{3pt}r{3pt}){1-6}
      StyleNet~\cite{128floats}     & 0.133 & 0.187 & 0.128 & 0.141 & 0.113\\
      Attributes      & 0.175~/~0.136 & 0.172~/~0.115 & 0.050~/~0.096 & 0.185~/~0.132 & 0.090~/~0.198\\
      PolyLDA  & 0.178~/~\bf0.303 & \textbf{0.424}~/~0.180 & 0.191~/~\bf0.266 & 0.130~/~\bf0.281 & 0.139~/~\bf0.394\\

   \end{tabular}\vspace{-2mm}
   \caption{Accuracy (AP) of retrieving a mixture of styles.}
   \label{tab:mix_style}
\end{table}

\section{Conclusion}

This work explores unsupervised discovery of complex styles in fashion.  We develop an approach based on polylingual topic models to model the composition of outfits from visual attributes.  The resulting styles offer a fine-grained representation valuable for organizing unlabeled fashion photos beyond their superficial visual ties (e.g., same literal garments or attributes).  While by necessity our results rely on external style labels for evaluation, we stress that the generality of the discovered styles is an asset, and they offer representational power beyond what traditional (supervised) classification schemes can do.  Our example results highlighting blended styles, trajectories between styles, and style summaries suggest a few such applications of interest.

\vspace*{-0.1in}
\paragraph{Acknowledgements:} We thank Suyog Jain and Chao-Yuan Wu for helpful discussions.  This research is supported in part by NSF IIS-1065390 and a gift from Amazon.

{\small
\bibliographystyle{ieee}
\bibliography{strings,egbib,refs}
}


\end{document}